\pdfoutput=1

\documentclass[11pt]{article}

\usepackage{EMNLP2023}
\usepackage{adjustbox}
\usepackage{booktabs}
\usepackage{tabularx}
\usepackage{times}
\usepackage{latexsym}
\usepackage{tabu}

\usepackage[T1]{fontenc}

\usepackage[utf8]{inputenc}

\usepackage{microtype}

\usepackage{inconsolata}

\title{ToddlerBERTa: Exploiting BabyBERTa for Grammar Learning and Language Understanding}

\author{Ömer Veysel Çağatan \\
    Koç University \\
   Rumelifeneri, Sarıyer Rumeli Feneri Yolu \\
   34450 Sarıyer/İstanbul,Turkey\\
  \texttt{ocagatan19@ku.edu.tr}}

\begin{document}
\maketitle

\begin{abstract}
We present ToddlerBERTa, a scaled BabyBERTa language model, exploring its capabilities through five different models with varied hyperparameters. We obtain our best model named ToddlerBERTa by meticulously optimizing our models on the BLiMP benchmark. Despite training on a smaller dataset, ToddlerBERTa demonstrates commendable performance, outperforming the baselines provided by a significant margin in the overall evaluation that includes BLiMP, SuperGLUE, MSGS and BLiMP supplement. ToddlerBERTa showcases robust language understanding, even with single-sentence pretraining, and competes with baselines that leverage broader contextual information. Our work provides insights into hyperparameter choices, and data utilization, contributing to the advancement of low-resource language models.\end{abstract}

\section{Introduction}
Over the past few years, there has been a lot of effort put into improving the pretraining of large language models (LLMs) on a large scale~\cite{Brown2020LanguageMA,Raffel2019ExploringTL,Chowdhery2022PaLMSL,Hoffmann2022TrainingCL}. While there is often a focus on increasing the number of parameters, there has also been significant growth in dataset size. However, there has been minimal progress in pretraining on smaller data scales that are comparable to how humans learn language.

Exploring pretraining on a smaller scale can serve as a trial area for developing original techniques that boost data effectiveness. These techniques can be scaled up to larger datasets utilized and employed to enhance current methods for modelling low-resource languages.

The BabyLM challenge~\cite{warstadt-et-al-2023-babylm} has been created to address the gap in research on pretraining for small-scale language models. Our focus will be on a limited corpus of approximately 10 million words, which includes child-directed speech, transcribed speech from various sources, children's books, and Wikipedia data.

We trained more than 180 BabyBERTa~\cite{Huebner2021BabyBERTaLM} models in different sizes and hyperparameters to determine how well language models learn grammar and understand language. Our findings showed that scaling the model and data resulted in significantly better outcomes compared to baseline models which underscores the low utilisation of both the data and architecture we currently have. All in all, our work demonstrates that well-known and widely used~\cite{Liu2019RoBERTaAR,Devlin2019BERTPO,Vaswani2017AttentionIA} architectures can be enhanced with moderate modifications to their training recipes.     
\section{Related Work}

There has been a significant amount of research on data-efficient language models. These models aim to achieve high accuracy in language tasks while using less training data than their larger counterparts. One way to create data-efficient language models is to reduce the number of model parameters while maintaining high performance. For instance, DistilBERT~\cite{Sanh2019DistilBERTAD} is a smaller and faster version of the popular BERT model. It was trained by distilling knowledge from the larger model into a smaller version. TinyBERT~\cite{Jiao2019TinyBERTDB}, on the other hand, was designed for low-resource environments, such as mobile devices. It was trained using a combination of teacher-student learning and knowledge distillation techniques.

Another example of a data-efficient language model is ALBERT~\cite{Lan2019ALBERTAL} which reduces the number of parameters of the BERT model by using factorization techniques and sharing parameters across different layers. This results in a more data-efficient model that can achieve similar or better performance than the larger BERT model.

GPT-Neo~~\cite{Black2021GPTNeoLS} is another data-efficient language model that was trained on a large dataset of text, but it can be fine-tuned on smaller datasets with good results. It has demonstrated competitive performance on various natural language processing tasks, including language generation, summarization, and question-answering.

\begin{table*}[htbp]
  \centering
  \caption{Model Configurations of ToddlerBERTa.}
  \label{table1}
  \begin{tabularx}{\textwidth}{llllll}
  \toprule
    & \textbf{Hidden Size} & \textbf{Inter. Size} & \textbf{\# Heads} & \textbf{\# Layers} &\textbf{\# Parameters} \\
    \midrule
    \textbf{ToddlerBERTa-xs} &  64 & 256 & 4 & 4 & 0.75 M \\
    \textbf{ToddlerBERTa-s} &  128 & 512 & 4 & 4 & 1.8 M \\
    \textbf{ToddlerBERTa-base} & 256 & 1024 & 8 & 8 & 8.5 M \\
   \textbf{ToddlerBERTa-l} &  512 & 2048 & 8 & 8 & 29.7 M \\
    \textbf{ToddlerBERTa-xl} &  768 & 3072 & 12 & 12 & 92.0 M \\
    \bottomrule
  \end{tabularx}
\end{table*}

ELECTRA~\cite{Clark2020ELECTRAPT} is a novel pre-training approach for language models that is designed to be more data-efficient than traditional models like BERT. Instead of using a traditional masked language modelling task, ELECTRA uses a discriminator network to predict whether a given input is real or generated by another model. This approach allows for more efficient training and can achieve similar or better performance than traditional models.

TinyStories~\cite{Eldan2023TinyStoriesHS} is an artificial collection of short stories, specifically designed with words understandable to 3 to 4-year-olds. These stories are generated using GPT-3.5 and GPT-4~\cite{OpenAI2023GPT4TR}.TinyStories can effectively serve as a training and evaluation dataset for language models (LMs) that are considerably smaller than the current state-of-the-art models (less than 10 million parameters) or have simpler architectures (with just one transformer block). Despite their reduced size and simplicity, these LMs are capable of producing coherent and consistent stories spanning multiple paragraphs. The stories are diverse, exhibit nearly flawless grammar, and showcase impressive reasoning abilities.

BabyBERTa is a lightweight model for language acquisition~\cite{Huebner2021BabyBERTaLM}. BabyBERTa is similar to RoBERTa~\cite{Liu2019RoBERTaAR}, but it is much smaller and simpler. BabyBERTa was trained on a dataset of 5M words of American-English child-directed input, and it can be run on a single desktop with a single GPU.BabyBERTa was able to achieve comparable performance to RoBERTa on a number of language acquisition tasks, including grammatical knowledge acquisition, generalization to novel grammatical contexts, syntactic structure learning, and semantic word and phrase learning. These results suggest that BabyBERTa could be a valuable tool for language acquisition research.

\textbf{Small size}: BabyBERTa is much smaller than RoBERTa, with only 8 layers, 8 attention heads, 256 hidden units, and an intermediate size of 1024. This makes it much faster and easier to train and use than RoBERTa.

\textbf{Comparable performance}: Despite its smaller size and simpler training regime, BabyBERTa was able to achieve comparable performance to RoBERTa on a number of language acquisition tasks. This suggests that BabyBERTa could be a valuable tool for language acquisition research.

BabyBERTa makes a number of contributions to the field. First, it demonstrates that a small, lightweight model can be used to acquire grammatical knowledge from child-directed input. Second, it shows that BabyBERTa can generalize to novel grammatical contexts. Third, it shows that BabyBERTa is able to learn the syntactic structure of sentences. Fourth, it shows that BabyBERTa is able to learn the semantics of words and phrases

\section{Experiment Settings}
We embrace BabyBERTa~\cite{Huebner2021BabyBERTaLM} as the foundational model for our research endeavour. Building upon this foundation, our investigation sets forth to explore an array of model sizes and diverse hyperparameters in a systematic and rigorous manner.

We construct five different models to validate and then further exploit the performance of BabyBERTa. All hyperparameters are kept the same except, hidden size, intermediate size, number of attention heads and number of layers. Models configurations can be found in Table \ref{table1}.

Our study closely follows the established hyperparameters of BabyBERTa but with three key variations: number of mask patterns\{1, 5, 10, 20, 50\}, epochs\{1,5,10\}, and batch size \{16,32,64,128\}. Due to computational limitations, we are limited to having 36 different configurations per model.

\section{Evaluation Setup}
We adopt the official evaluation pipeline of the BabyLM Challenge~\cite{warstadt-et-al-2023-babylm,eval-harness}, which combines BLiMP~\cite{Warstadt2019BLiMPAB}, SuperGLUE~\cite{Wang2019SuperGLUEAS}, MSGS~\cite{Warstadt2020LearningWF}, and a Supplement benchmark. Our best model is evaluated on all benchmarks, while other models are evaluated on BLiMP due to limited computing resources. This approach ensures a rigorous assessment of our model's performance across diverse tasks while optimizing resource allocation.

\subsection{Baselines}
The competition organizers supply baseline models extracted from well-known language models, including OPT~\cite{Zhang2022OPTOP}, RoBERTa~\cite{Liu2019RoBERTaAR}, and T5~\cite{Raffel2019ExploringTL}. These baselines are trained from scratch on the competition's exclusive dataset. Since no external models are available, we use these baseline models as references to assess our models' performance within the competition's context.
\section{Results and Analysis}
As stipulated earlier, a substantial portion of our model evaluations is conducted under BLiMP~\cite{Warstadt2019BLiMPAB}, encompassing comparisons across various linguistic tasks. Additionally, we undertake a comprehensive evaluation of our best-performing model using the entire prescribed evaluation pipeline. As a result, we present our findings as two distinct sets of results: BLiMP results and main results.

\begin{figure}[htbp]
  \centering
  \includegraphics[width=0.5\textwidth]{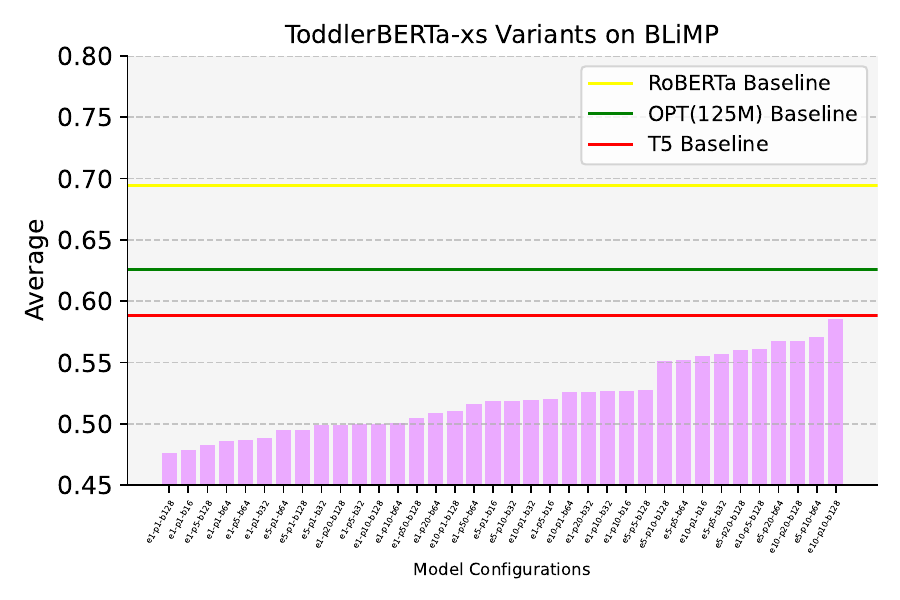} 
  \caption{Average scores of the ToddlerBERTa-xs models on BLiMP are reported. We shorten the different configuration names as number of epochs: e, number of dynamic patterns: p and batch size: b.}
  \label{fig:plt1}
\end{figure}
\subsection{BliMP Results}
\subsubsection{ToddlerBERTa-xs}
Our ToddlerBERTa-xs model, with approximately 750 thousand parameters, achieves competitive performance compared to the larger T5 baseline on the BLiMP benchmark, in Figure \ref{fig:plt1}. This data scaling behaviour highlights the potential benefits of optimizing smaller architectures for specific tasks, showcasing efficient language modelling approaches.

\subsubsection{ToddlerBERTa-s}
ToddlerBERTa-s model, consisting of 1.8 million parameters, exhibits superior performance compared to the OPT baseline across various configurations. Remarkably, experimental results demonstrate that even with smaller parameter sizes, these models can outperform larger counterparts in the low data regime when leveraging the BabyBERTa training and preprocessing recipes.
\begin{figure}[htbp]
  \centering
  \includegraphics[width=0.5\textwidth]{
  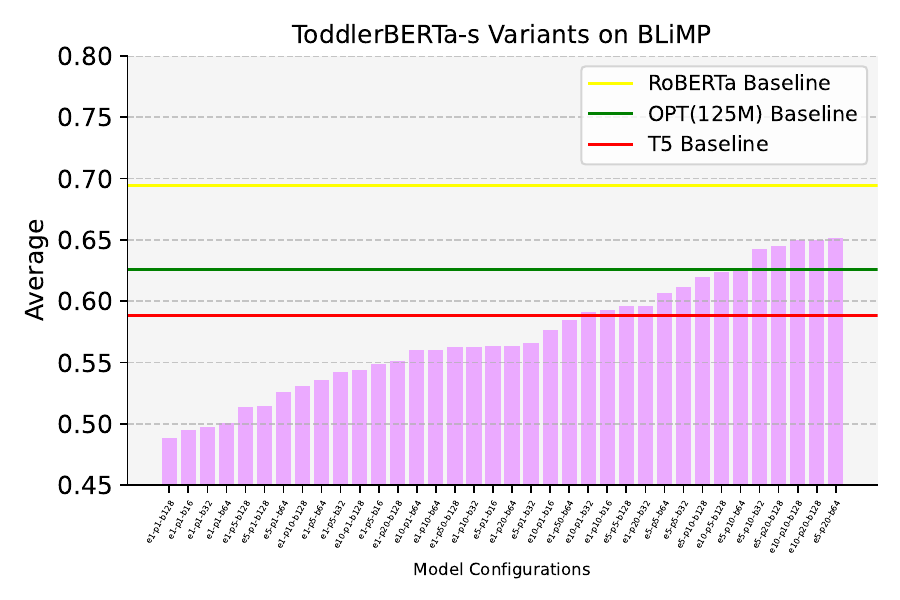} 
  \caption{Average scores of the ToddlerBERTa-s models on BLiMP are reported. We shorten the different configuration names as number of epochs: e, number of dynamic patterns: p and batch size: b.}
  \label{fig:plt2}
\end{figure}

\subsubsection{ToddlerBERTa-base}
\begin{figure}[htbp]
  \centering
  \includegraphics[width=0.5\textwidth]{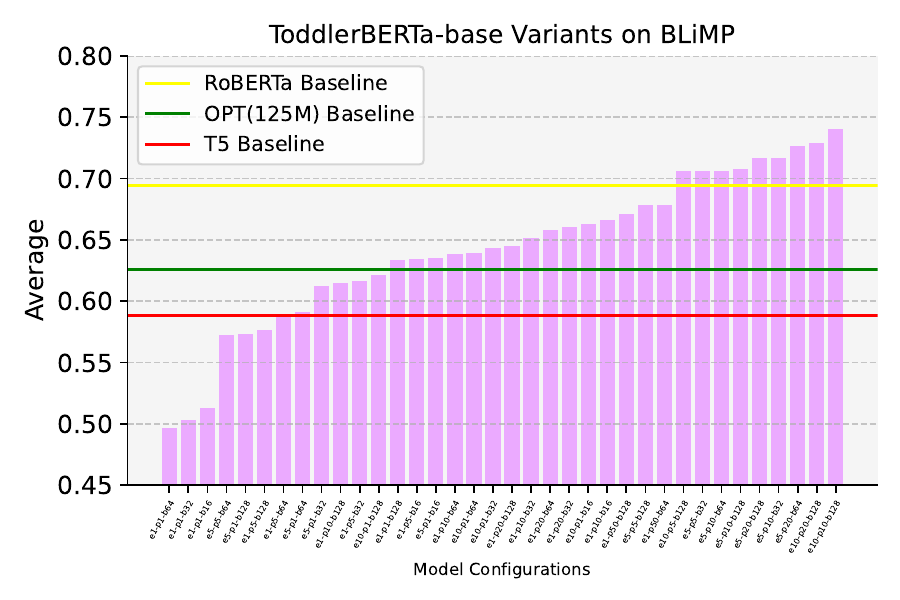} 
  \caption{Average scores of the ToddlerBERTa-base models on BLiMP are reported. We shorten the different configuration names as number of epochs: e, number of dynamic patterns: p and batch size: b.}
  \label{fig:plt3}
\end{figure}

The ToddlerBERTa-base and BabyBERTa~\cite{Huebner2021BabyBERTaLM} have the same number of parameters, which is 8.5 million. However, the best-performing model of ToddlerBERTa-base scores 0.7407 with more epochs and mask patterns than the original, as shown in Figure \ref{fig:plt3}. On the other hand, the original BabyBERTa~\cite{Huebner2021BabyBERTaLM} configuration achieves 0.6660.

\subsubsection{ToddlerBERTa-l}

\begin{figure}[htbp]
  \centering
  \includegraphics[width=0.5\textwidth]{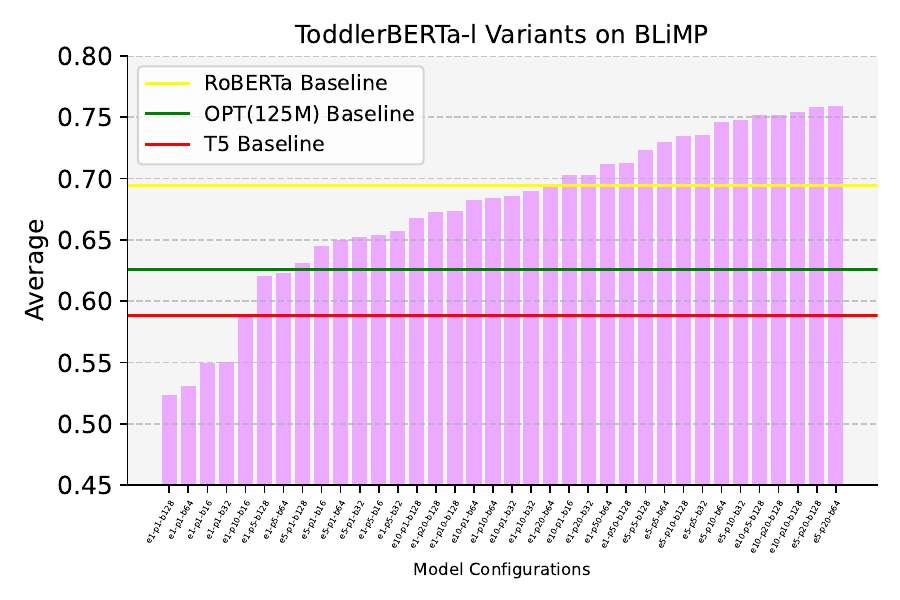} 
  \caption{Average scores of the ToddlerBERTa-l models on BLiMP are reported. We shorten the different configuration names as number of epochs: e, number of dynamic patterns: p and batch size: b.}
  \label{fig:plt4}
\end{figure}

The utilization of data scaling techniques is evidently advantageous in enhancing model performance for grammar learning tasks. However, our research findings demonstrate that surpassing the RoBERTa baseline is achievable through the increase of model parameters. This observation prompts an inquiry into the sustainability of this trend. In order to address this question, we developed ToddlerBERTa-l, featuring a substantial parameter count of approximately 30 million. Our experimental results emphasize the indispensability of model size, despite the relatively modest increase in the top score, Figure \ref{fig:plt4}. Notably, a significant performance boost is observed in the majority of models when larger architectures are employed. These findings underscore the critical role of model size in optimizing grammar learning capabilities.

\subsubsection{ToddlerBERTa-xl}

\begin{figure}[htbp]
  \centering
  \includegraphics[width=0.5\textwidth]{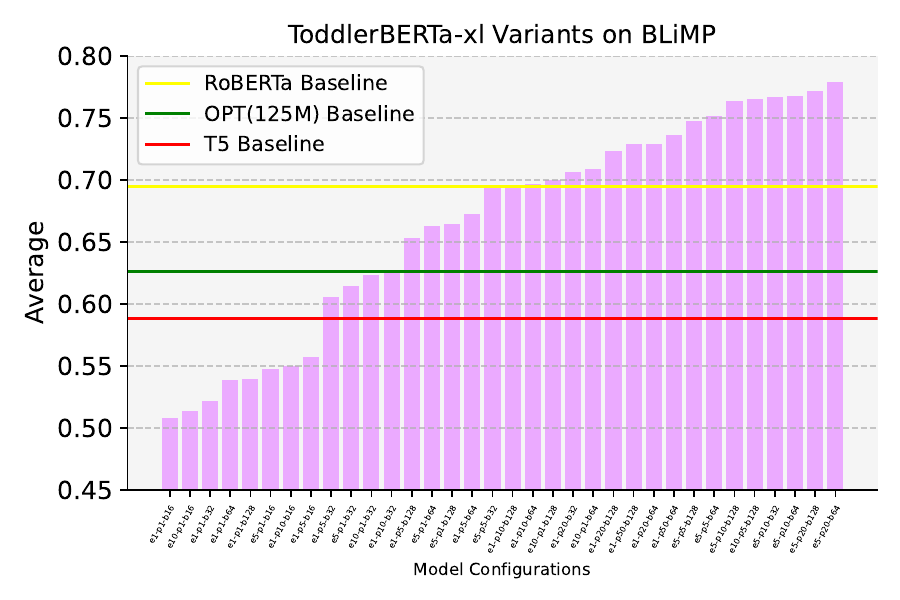} 
  \caption{Average scores of the ToddlerBERTa-xl models on BLiMP are reported. We shorten the different configuration names as number of epochs: e, number of dynamic patterns: p and batch size: b.}
  \label{fig:plt5}
\end{figure}

To further explore the capabilities of BabyBERTa within the strict-small portion of BabyLM, we introduce ToddlerBERTa-xl, a language model equipped with 92 million parameters similar to RoBERTa~\cite{Liu2019RoBERTaAR}. Our prior experiments have highlighted the significance of both data and model size; however, these studies have predominantly employed relatively smaller model sizes compared to baseline models, which exhibit exceptional results when trained on extended corpora over extended periods. Such large models excel under substantial data volumes but tend to perform inadequately in low-data scenarios. Consequently, previous investigations \cite{Eldan2023TinyStoriesHS, Huebner2021BabyBERTaLM} have often opted for smaller model sizes. Nonetheless, to thoroughly evaluate the boundaries of this approach, we undertake the training of larger models in order to affirm our hypothesis which is that performance will improve with the model scaling. Figure \ref{fig:plt5} verifies our hypothesis by achieving remarkable results on BLiMP with a significant margin to baselines which share a similar number of parameters.

\newcommand\rotation{30} 
\begin{table*}[ht]
\begin{minipage}{1.0\linewidth}
	\centering
	\footnotesize
	\addtolength{\tabcolsep}{-3.8pt}

\begin{tabu}{@{}ll@{\hskip 8pt}llllllllllll@{\hskip 8pt}lll@{}}

\multicolumn{1}{p{2.5ex}}{\rotatebox{0}{\textbf{Models}}} &
\multicolumn{1}{p{2.5ex}}{\rotatebox{\rotation}{\textbf{Overall}}} &
\multicolumn{1}{p{2.5ex}}{\rotatebox{\rotation}{\textsc{Ana.\ agr}}} &
\multicolumn{1}{p{2.5ex}}{\rotatebox{\rotation}{\textsc{Arg.\ str}}} &
\multicolumn{1}{p{2.5ex}}{\rotatebox{\rotation}{\textsc{Binding}}} &
\multicolumn{1}{p{2.5ex}}{\rotatebox{\rotation}{\textsc{Ctrl.\ rais.}}} &
\multicolumn{1}{p{2.5ex}}{\rotatebox{\rotation}{\textsc{D-n agr}}} &
\multicolumn{1}{p{2.5ex}}{\rotatebox{\rotation}{\textsc{Ellipsis}}} &
\multicolumn{1}{p{2.5ex}}{\rotatebox{\rotation}{\textsc{Filler\ gap}}} &
\multicolumn{1}{p{2.5ex}}{\rotatebox{\rotation}{\textsc{Irregular}}} &
\multicolumn{1}{p{2.5ex}}{\rotatebox{\rotation}{\textsc{Island}}} &
\multicolumn{1}{p{2.5ex}}{\rotatebox{\rotation}{\textsc{NPI}}} &
\multicolumn{1}{p{2.5ex}}{\rotatebox{\rotation}{\textsc{Quantifiers}}} &
\multicolumn{1}{p{2.5ex}}{\rotatebox{\rotation}{\textsc{S-v agr}}} &
 \\
\toprule 

OPT-125m(baseline) & \textbf{62.63} & 63.75 & 70.56 & 67.10 & 66.48 & 78.47 & 62.01 & 63.83 & 67.53 & 48.58 & 46.71 & 59.61 & 56.87  \\

RoBERTa-base(baseline) & \textbf{69.47} & 81.54 & 67.12 & 67.26 & 67.85 & 90.75 & 76.44 & 63.48 & 87.43 & 39.87 & 55.92 & 70.53 & 65.42  \\
T5(baseline) & \textbf{57.70} & 68.92 & 63.82 & 60.40 & 60.87 & 72.21 & 34.41 & 48.24 & 77.56 & 45.59 & 47.80 & 56.72 & 55.81 \\
ToddlerBERTa & \textbf{76.68} & 87.68 & 70.62 & 71.82 & 69.07 & 93.44 & 76.27 & 81.68 & 82.80  & 58.07 & 63.59 & 82.64 & 82.51 \\
\midrule
Roberta-base & \textbf{85.4} & 97.30 & 83.50 & 77.80 & 81.9 & 97.00 & 91.40 & 90.10 & 96.20 & 80.70 & 81.00 & 69.80 & 91.90 \\

\bottomrule
\end{tabu}
\end{minipage}
\caption{BLiMP\cite{Warstadt2019BLiMPAB} benchmark results, baseline scores are taken from the  \href{https://dynabench.org/tasks/baby_strict_small}{leaderboard} page of the competition , RoBERTa-base results from \cite{Huebner2021BabyBERTaLM}.}
\label{table:accuracy-blimp}
\end{table*}

\begin{table*}[ht]
\begin{minipage}{1.0\linewidth}
	\centering
	\footnotesize
	\addtolength{\tabcolsep}{-3.8pt}

\begin{tabu}{@{}ll@{\hskip 8pt}llllll@{\hskip 8pt}lll@{}}
\multicolumn{1}{p{2.5ex}}{\rotatebox{0}{\textbf{Models}}} &
\multicolumn{1}{p{2.5ex}}{\rotatebox{\rotation}{\textbf{Overall}}} &
\multicolumn{1}{p{2.5ex}}{\rotatebox{\rotation}{\textsc{Hypernym}}} &
\multicolumn{1}{p{2.5ex}}{\rotatebox{\rotation}{\textsc{QA Congr.(easy)}}} &
\multicolumn{1}{p{2.5ex}}{\rotatebox{\rotation}{\textsc{QA Congr.(tricky)}}} &
\multicolumn{1}{p{2.5ex}}{\rotatebox{\rotation}{\textsc{Subj.-Aux. Inver.}}} &
\multicolumn{1}{p{2.5ex}}{\rotatebox{\rotation}{\textsc{Turn Taking}}} &

 &
 \\
\toprule 
OPT-125m(baseline) & \textbf{52.72} & 50.00 & 54.69 & 31.52 & 70.26 & 57.14 \\
\
RoBERTa-base(baseline) & \textbf{42.42} & 50.80 & 34.40 & 34.50	& 45.60 & 46.80 \\
T5(baseline) & \textbf{43.96} & 48.02 & 40.63  & 21.21 & 64.92	& 45.00  \\

ToddlerBERTa & \textbf{57.12}& 48.02 & 62.50 & 35.76  & 79.65 & 59.64   \\

\end{tabu}
\end{minipage}
\caption{BLiMP Supplement benchmark results, baseline scores are taken from the GitHub page of \href{https://github.com/babylm/evaluation-pipeline}{evaluation pipeline}.}
\label{table:accuracy-blimp-supp}
\end{table*}
\subsubsection{BLiMP Summary}
Our extensive experiments show that improving the BabyBERTa methodology involves using numerous different mask patterns to augment the data, processing single sentences, and using smaller context and vocabulary sizes with limited batch sizes and epochs. However, to achieve superior performance with larger models, we increase batch sizes and the number of epochs. Larger batch sizes enhance training stability, while more epochs help models learn better. Consequently, our best model outperforms the original BabyBERTa model by a substantial 10 points in BLiMP, highlighting the effectiveness of these changes.
\begin{figure}[htbp]
  \centering
  \includegraphics[width=0.6\textwidth]{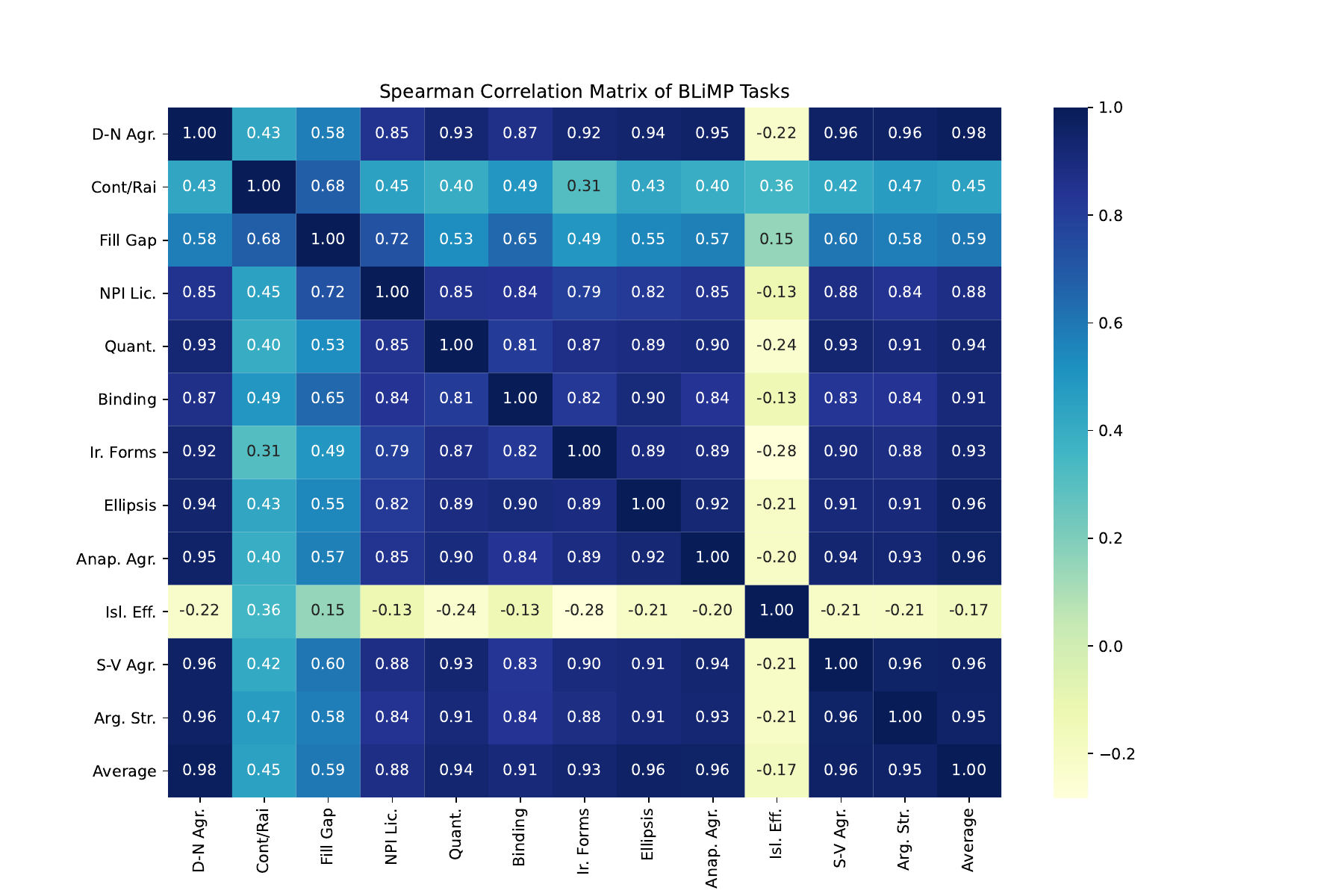} 
  \caption{Spearman correlation matrix on the scores of BLiMP tasks.}
  \label{fig:heatmap}
\end{figure}

To refine our models based on BLiMP evaluation, we carefully consider the average results while remaining aware of potential outliers that could have an implicit impact on the reliability of the approach that we take while optimizing the models. To thoroughly explore relationships among the nearly 180 results of our models, we use a Spearman correlation matrix as a robust analytical tool, providing insights into potential patterns and dependencies. See Figure \ref{fig:heatmap} for the correlation matrix

\begin{figure}[htbp]
  \centering
  \includegraphics[width=0.45\textwidth]{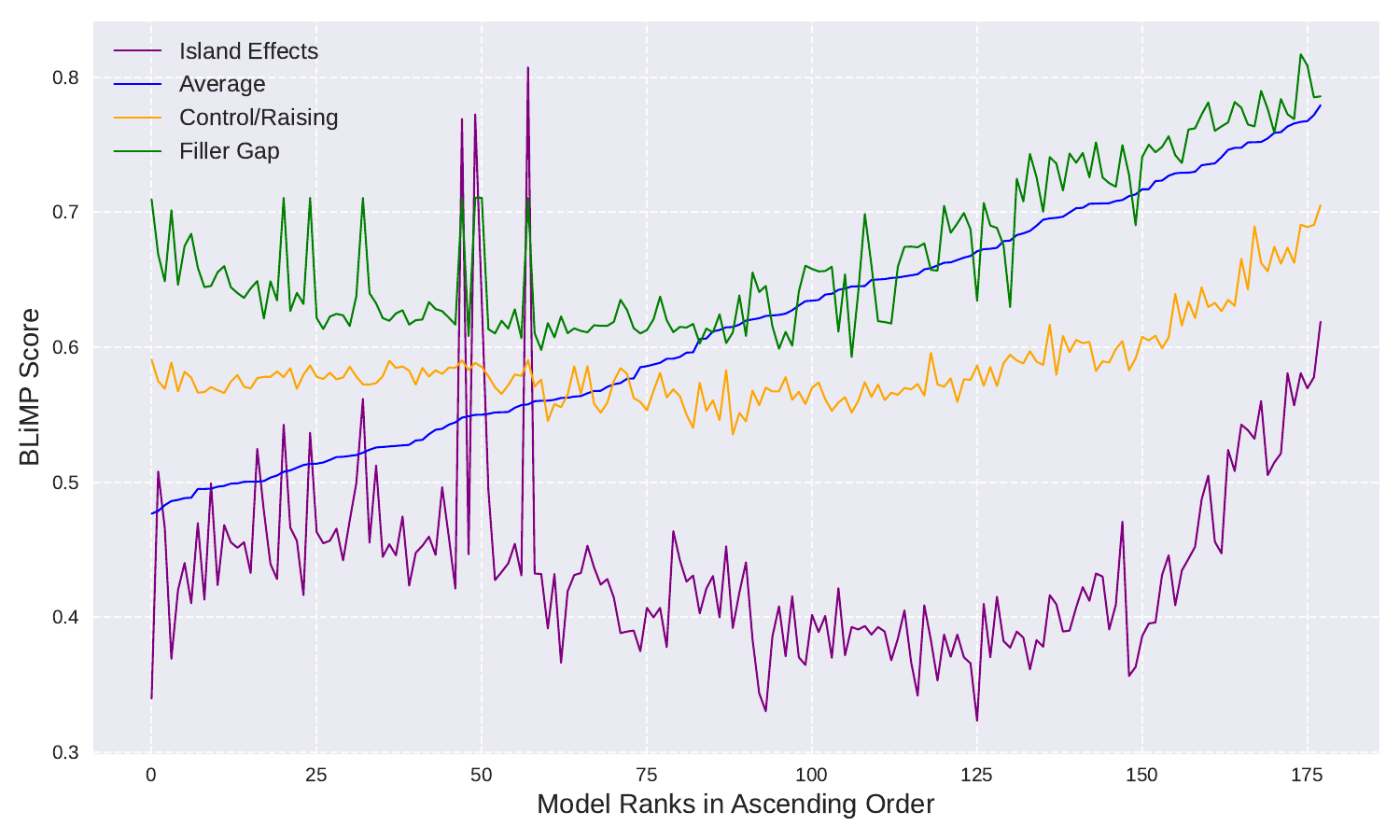} 
  \caption{Models are ranked by the average BLiMP score in ascending order, in the Blue time series plot. Other time series plots represent how task scores vary while the average score consistently improves.}
  \label{fig:time}
\end{figure}
\begin{table*}[ht]
\begin{minipage}{1\linewidth}
	\centering
	\footnotesize
	\addtolength{\tabcolsep}{-3.8pt}

\begin{tabu}{@{}ll@{\hskip 6pt}llllllllllll@{\hskip 8pt}lll@{}}
\toprule 

\multicolumn{1}{p{2.5ex}}{\rotatebox{0}{\textbf{Models}}} &
\multicolumn{1}{p{2.5ex}}{\rotatebox{\rotation}{\textbf{Overall}}} &
\multicolumn{1}{p{2.5ex}}{\rotatebox{\rotation}{\textsc{CR}}} &
\multicolumn{1}{p{2.5ex}}{\rotatebox{\rotation}{\textsc{LC}}} &
\multicolumn{1}{p{2.5ex}}{\rotatebox{\rotation}{\textsc{MV}}} &
\multicolumn{1}{p{2.5ex}}{\rotatebox{\rotation}{\textsc{RP}}} &
\multicolumn{1}{p{2.5ex}}{\rotatebox{\rotation}{\textsc{SC}}} &
\multicolumn{1}{p{2.5ex}}{\rotatebox{\rotation}{\textsc{CR\_LC}}} &
\multicolumn{1}{p{2.5ex}}{\rotatebox{\rotation}{\textsc{CR\_RTP}}} &
\multicolumn{1}{p{2.5ex}}{\rotatebox{\rotation}{\textsc{MV\_LC}}} &
\multicolumn{1}{p{2.5ex}}{\rotatebox{\rotation}{\textsc{MV\_RTP}}} &
\multicolumn{1}{p{2.5ex}}{\rotatebox{\rotation}{\textsc{SC\_LC}}} &
\multicolumn{1}{p{2.5ex}}{\rotatebox{\rotation}{\textsc{SC\_RP}}} &
 &
 \\
\midrule 
OPT-125m(baseline) & \textbf{9.63} & 50.77 & 53.55 & 99.47 & 99.91 & 77.15 & 0.37 & -70.33 & -72.14 & -77.60 & 13.76 & -68.92  \\
RoBERTa-base(baseline) & \textbf{8.22} & 43.08 & 100.00  & 97.67 & 76.73 & 86.24 & -28.28 & -77.69 & -99.30 & -79.36 & 16.28 & -45.02 \\
T5(baseline) & \textbf{-6.38} & 21.11 & 100.00& 33.36 & 82.54 & 77.58 & -78.33 & -62.04 & -100.00 & -79.70 & -25.28 & -39.43 \\

ToddlerBERTa & \textbf{2.51} & 51.61 & 80.00 & 99.95 & 71.23 & 45.90 & 2.32 & -72.15 & -85.73 & -82.68 & -34.41 & -49.60  \\
\bottomrule

\end{tabu}
\end{minipage}
\caption{MSGS~\cite{Warstadt2020LearningWF} benchmark results, baseline scores are taken from the GitHub page of \href{https://github.com/babylm/evaluation-pipeline}{evaluation pipeline}}
\label{table:msgs}
\end{table*}

\begin{table*}[ht]
\begin{minipage}{1.0\linewidth}
	\centering
	\footnotesize
	\addtolength{\tabcolsep}{-3.8pt}

\begin{tabu}{@{}ll@{\hskip 8pt}llllllllllll@{\hskip 8pt}lll@{}}
\multicolumn{1}{p{2.5ex}}{\rotatebox{0}{\textbf{Models}}} &
\multicolumn{1}{p{2.5ex}}{\rotatebox{\rotation}{\textbf{Overall}}} &
\multicolumn{1}{p{2.5ex}}{\rotatebox{\rotation}{\textsc{CoLA(MCC)}}} &
\multicolumn{1}{p{2.5ex}}{\rotatebox{\rotation}{\textsc{SST-2}}} &
\multicolumn{1}{p{2.5ex}}{\rotatebox{\rotation}{\textsc{MRPC(F1)}}} &
\multicolumn{1}{p{2.5ex}}{\rotatebox{\rotation}{\textsc{QQP(F1)}}} &
\multicolumn{1}{p{2.5ex}}{\rotatebox{\rotation}{\textsc{MNLI}}} &
\multicolumn{1}{p{2.5ex}}{\rotatebox{\rotation}{\textsc{MNLI-mm}}} &
\multicolumn{1}{p{2.5ex}}{\rotatebox{\rotation}{\textsc{QNLI}}} &
\multicolumn{1}{p{2.5ex}}{\rotatebox{\rotation}{\textsc{RTE}}} &
\multicolumn{1}{p{2.5ex}}{\rotatebox{\rotation}{\textsc{BoolQ}}} &
\multicolumn{1}{p{2.5ex}}{\rotatebox{\rotation}{\textsc{MultiRC}}} &
\multicolumn{1}{p{2.5ex}}{\rotatebox{\rotation}{\textsc{WSC}}} &
 &
 \\
\toprule 
OPT-125m(baseline) & \textbf{62.38} & 15.22 & 84.25 & 74.13 & 78.89 & 67.66 & 69.43 & 65.40 & 55.26 & 65.28 & 51.37 & 59.04  \\
RoBERTa-base(baseline) & \textbf{67.38} & 25.75 & 87.60 & 77.27	& 82.76 & 73.15 & 77.27 & 81.54 & 53.54 & 65.70	& 61.23 & 57.83 \\
T5(baseline) & \textbf{58.34} & 11.26 & 80.91 & 78.49 & 72.19 & 52.80 & 56.70 & 63.91 & 50.51 & 63.49 & 48.85	& 62.65 \\

ToddlerBERTa & \textbf{64.94} & 37.37 & 86.02 & 79.29 & 74.53 & 70.28 & 70.34 & 64.83 & 54.55 & 67.77 & 47.97 & 61.45   \\
\bottomrule

\end{tabu}
\end{minipage}
\caption{SuperGLUE~\cite{Wang2019SuperGLUEAS} benchmark results, baseline scores are taken from the GitHub page of \href{https://github.com/babylm/evaluation-pipeline}{evaluation pipeline}}
\label{table:glue}
\end{table*}

The majority of the tasks exhibit a strong positive correlation with the average, with the exception of Island Effects, Filler Gap, and Control/Raising. In order to gain insights into the underlying reasons behind this anomaly, we present a visual analysis by plotting the scores of these specific tasks in ascending order based on their respective average scores, as illustrated in Figure \ref{fig:time}. The plot reveals that all task scores either improve slightly or stay around a fixed interval. This observation leads us to postulate that these particular tasks may be inherently more challenging, demanding a larger volume of data and more complex model architectures for optimal performance.

\subsection{Main Results}

After evaluating various models on BLiMP~\cite{Warstadt2019BLiMPAB}, we select the best one as our final model which is a ToddlerBERTa-xl that is trained for 5 epochs with 20 different mask patterns and 64 as the batch size. We then assess its performance on Blimp Supplement and fine-tune it on~\cite{Wang2019SuperGLUEAS} and MSGS~\cite{Warstadt2020LearningWF} using the evaluation pipeline~\cite{warstadt-et-al-2023-babylm}.

\textbf{BLiMP}: In our investigation, we focus on evaluating our models compared to baselines during iterative training. We also include results of RoBERTa-base~\cite{Liu2019RoBERTaAR} from ~\citet{Huebner2021BabyBERTaLM} for a more comprehensive analysis in Table \ref{table:accuracy-blimp}. RoBERTa-base outperforms our ToddlerBERTa model, largely due to its extensive 3-billion-word training data, while ToddlerBERTa is trained on a smaller 10-million-word dataset.

To narrow the performance gap, we increase mask patterns in ToddlerBERTa's training, improving data utilization despite the 1-billion-word exposure constraint. Our results show that ToddlerBERTa, with limited data, can perform relatively well compared to RoBERTa-base, highlighting the effectiveness of data augmentation by employing different masks for enhancing language model training.

\textbf{SuperGLUE}: In the SuperGLUE benchmark, our models face a challenge due to their exclusive focus on single sentences while the dataset often includes inputs with multiple sentences. However, even with this constraint, our model competes remarkably well with baselines trained on multiple sentences. Our results in Table \ref{table:glue}, highlight our model's ability to grasp complex linguistic relationships and reasoning, aligning its performance with state-of-the-art baselines that use broader contextual information. This showcases our model's potential for robust language understanding, even in scenarios with multi-sentence inputs.

\textbf{MSGS}: The Mixed Signals Generalization Set (MSGS) evaluates language models' generalization capabilities for both linguistic and surface features. Our analysis in Table \ref{table:msgs} suggests that the poor performance may be due in part to overexposure. To enhance training, we add more mask patterns and use them for numerous epochs, which can lead to repeated patterns and examples in the training data. This overexposure may affect the model's learning process, causing a preference for specific features. As a result, the model might struggle to adapt to novel patterns in the MSGS. On the other hand, baseline models also suffer from poor performance. Considering the worst score is -100 and the best is 100, their performances are no better than ours which points out that undertraining is another drawback for generalization.

\textbf{BLiMP Supplement}: The challenge has been enriched with an extra benchmark, the details of which have not been published yet, but it is presumed to be connected to the BLiMP evaluation framework. Analysis of the results presented in Table \ref{table:accuracy-blimp-supp} leads us to speculate that the performance gains in BLiMP are still relevant whereas insufficient to truly accomplish a major performance. ToddlerBERTa achieves better scores than the baselines however performance of OPT-125m~\cite{Zhang2022OPTOP} and T5~\cite{Raffel2019ExploringTL} compared to RoBERTa~\cite{Liu2019RoBERTaAR} can be explained by the presence of the decoder in T5 and OPT architectures. Further analysis will be ineffective given that details of benchmark are non-disclosed yet.
\section{Conclusion}

We undertake a systematic and rigorous exploration of language models, building upon the foundational work of BabyBERTa. Through the development and evaluation of five distinct ToddlerBERTa models, we have demonstrated the significance of hyperparameter choices and model sizes in the context of natural language processing.

Our experiments have revealed the potential benefits of optimizing smaller architectures for specific linguistic tasks, showcasing the efficiency of language modelling techniques in tackling various challenges. Additionally, our best-performing ToddlerBERTa models have exhibited competitive performance compared to established baselines, showcasing their adaptability and capacity to excel in diverse language understanding tasks.

The comprehensive evaluations conducted on BLiMP, SuperGLUE, MSGS, and the new BLiMP Supplement benchmark have provided valuable insights into the strengths and limitations of our approach. While our research has shed light on the impact of different hyperparameters, we acknowledge that further exploration of model architectures and training methodologies may yield additional advancements in language modelling.

By contributing to the collective understanding of transformer-based models and their potential for natural language processing, our research aims to inspire future investigations and innovations in the field. As the quest for advancements in language modelling continues, we emphasize the importance of replicability and reproducibility in research to facilitate the development of robust and reliable language models.

\section{Limitations}
Despite the contributions of our research, it is essential to acknowledge its limitations. Firstly, the exploration of hyperparameters and model sizes may not have encompassed all possible configurations due to computational constraints. This leaves room for potential superior settings to be uncovered. Secondly, the evaluation framework's focus on transformer-based models may limit the comparability with other non-transformer architectures. Additionally, the fixed dataset used for training and evaluation may restrict the model's exposure to diverse linguistic patterns and contexts. Furthermore, the reliance on single-sentence processing during pretraining could impact the model's performance on tasks requiring broader contextual understanding. Lastly, our study did not extensively explore architectural innovations or novel training methodologies. Despite these limitations, our research provides valuable insights into language modelling, calling for further investigations to address these constraints and advance the field.

\section*{Ethics Statement}
The model under consideration, ToddlerBERTa, is devoid of generative capabilities, thereby ensuring that it cannot engender unfair, biased, or harmful content. The datasets employed in this study have been sourced from widely acknowledged repositories with an established reputation for safety in research applications, being meticulously selected to preclude the inclusion of personal information or offensive material.

\section*{Acknowledgements}
We would like to express our gratitude to the KUIS AI Center for their generous provision of computing resources for this project. We would also like to extend our appreciation to Gözde Gül Şahin for her valuable feedback and insightful discussions. 
\section*{Implementation and Hardware Details}
We use the \href{https://github.com/phueb/BabyBERTa}{official repository} of the BabyBERTa~\cite{Huebner2021BabyBERTaLM}. We use the transformers~\cite{Wolf2019HuggingFacesTS} to train our tokenizer and host our \href{https://huggingface.co/asparius/ToddlerBERTa}{best model}. We use the Tesla T4 and Tesla A100 provided by KUIS AI Center.

\bibliography{emnlp2023}

\begin{thebibliography}{22}
\expandafter\ifx\csname natexlab\endcsname\relax\def\natexlab#1{#1}\fi

\bibitem[{Black et~al.(2021)Black, Gao, Wang, Leahy, and
  Biderman}]{Black2021GPTNeoLS}
Sid Black, Leo Gao, Phil Wang, Connor Leahy, and Stella~Rose Biderman. 2021.
\newblock Gpt-neo: Large scale autoregressive language modeling with
  mesh-tensorflow.

\bibitem[{Brown et~al.(2020)Brown, Mann, Ryder, Subbiah, Kaplan, Dhariwal,
  Neelakantan, Shyam, Sastry, Askell, Agarwal, Herbert-Voss, Krueger, Henighan,
  Child, Ramesh, Ziegler, Wu, Winter, Hesse, Chen, Sigler, Litwin, Gray, Chess,
  Clark, Berner, McCandlish, Radford, Sutskever, and
  Amodei}]{Brown2020LanguageMA}
Tom~B. Brown, Benjamin Mann, Nick Ryder, Melanie Subbiah, Jared Kaplan,
  Prafulla Dhariwal, Arvind Neelakantan, Pranav Shyam, Girish Sastry, Amanda
  Askell, Sandhini Agarwal, Ariel Herbert-Voss, Gretchen Krueger, T.~J.
  Henighan, Rewon Child, Aditya Ramesh, Daniel~M. Ziegler, Jeff Wu, Clemens
  Winter, Christopher Hesse, Mark Chen, Eric Sigler, Mateusz Litwin, Scott
  Gray, Benjamin Chess, Jack Clark, Christopher Berner, Sam McCandlish, Alec
  Radford, Ilya Sutskever, and Dario Amodei. 2020.
\newblock Language models are few-shot learners.
\newblock \emph{ArXiv}, abs/2005.14165.

\bibitem[{Chowdhery et~al.(2022)Chowdhery, Narang, Devlin, Bosma, Mishra,
  Roberts, Barham, Chung, Sutton, Gehrmann, Schuh, Shi, Tsvyashchenko, Maynez,
  Rao, Barnes, Tay, Shazeer, Prabhakaran, Reif, Du, Hutchinson, Pope, Bradbury,
  Austin, Isard, Gur-Ari, Yin, Duke, Levskaya, Ghemawat, Dev, Michalewski,
  Garc{\'i}a, Misra, Robinson, Fedus, Zhou, Ippolito, Luan, Lim, Zoph,
  Spiridonov, Sepassi, Dohan, Agrawal, Omernick, Dai, Pillai, Pellat,
  Lewkowycz, Moreira, Child, Polozov, Lee, Zhou, Wang, Saeta, D{\'i}az, Firat,
  Catasta, Wei, Meier-Hellstern, Eck, Dean, Petrov, and
  Fiedel}]{Chowdhery2022PaLMSL}
Aakanksha Chowdhery, Sharan Narang, Jacob Devlin, Maarten Bosma, Gaurav Mishra,
  Adam Roberts, Paul Barham, Hyung~Won Chung, Charles Sutton, Sebastian
  Gehrmann, Parker Schuh, Kensen Shi, Sasha Tsvyashchenko, Joshua Maynez,
  Abhishek Rao, Parker Barnes, Yi~Tay, Noam~M. Shazeer, Vinodkumar Prabhakaran,
  Emily Reif, Nan Du, Benton~C. Hutchinson, Reiner Pope, James Bradbury, Jacob
  Austin, Michael Isard, Guy Gur-Ari, Pengcheng Yin, Toju Duke, Anselm
  Levskaya, Sanjay Ghemawat, Sunipa Dev, Henryk Michalewski, Xavier Garc{\'i}a,
  Vedant Misra, Kevin Robinson, Liam Fedus, Denny Zhou, Daphne Ippolito, David
  Luan, Hyeontaek Lim, Barret Zoph, Alexander Spiridonov, Ryan Sepassi, David
  Dohan, Shivani Agrawal, Mark Omernick, Andrew~M. Dai,
  Thanumalayan~Sankaranarayana Pillai, Marie Pellat, Aitor Lewkowycz, Erica
  Moreira, Rewon Child, Oleksandr Polozov, Katherine Lee, Zongwei Zhou, Xuezhi
  Wang, Brennan Saeta, Mark D{\'i}az, Orhan Firat, Michele Catasta, Jason Wei,
  Kathleen~S. Meier-Hellstern, Douglas Eck, Jeff Dean, Slav Petrov, and Noah
  Fiedel. 2022.
\newblock Palm: Scaling language modeling with pathways.
\newblock \emph{ArXiv}, abs/2204.02311.

\bibitem[{Clark et~al.(2020)Clark, Luong, Le, and Manning}]{Clark2020ELECTRAPT}
Kevin Clark, Minh-Thang Luong, Quoc~V. Le, and Christopher~D. Manning. 2020.
\newblock Electra: Pre-training text encoders as discriminators rather than
  generators.
\newblock \emph{ArXiv}, abs/2003.10555.

\bibitem[{Devlin et~al.(2019)Devlin, Chang, Lee, and
  Toutanova}]{Devlin2019BERTPO}
Jacob Devlin, Ming-Wei Chang, Kenton Lee, and Kristina Toutanova. 2019.
\newblock Bert: Pre-training of deep bidirectional transformers for language
  understanding.
\newblock \emph{ArXiv}, abs/1810.04805.

\bibitem[{Eldan and Li(2023)}]{Eldan2023TinyStoriesHS}
Ronen Eldan and Yuan-Fang Li. 2023.
\newblock \href {https://api.semanticscholar.org/CorpusID:258686446}
  {Tinystories: How small can language models be and still speak coherent
  english?}
\newblock \emph{ArXiv}, abs/2305.07759.

\bibitem[{Gao et~al.(2021)Gao, Tow, Biderman, Black, DiPofi, Foster, Golding,
  Hsu, McDonell, Muennighoff, Phang, Reynolds, Tang, Thite, Wang, Wang, and
  Zou}]{eval-harness}
Leo Gao, Jonathan Tow, Stella Biderman, Sid Black, Anthony DiPofi, Charles
  Foster, Laurence Golding, Jeffrey Hsu, Kyle McDonell, Niklas Muennighoff,
  Jason Phang, Laria Reynolds, Eric Tang, Anish Thite, Ben Wang, Kevin Wang,
  and Andy Zou. 2021.
\newblock \href {https://doi.org/10.5281/zenodo.5371628} {A framework for
  few-shot language model evaluation}.

\bibitem[{Hoffmann et~al.(2022)Hoffmann, Borgeaud, Mensch, Buchatskaya, Cai,
  Rutherford, de~Las~Casas, Hendricks, Welbl, Clark, Hennigan, Noland,
  Millican, van~den Driessche, Damoc, Guy, Osindero, Simonyan, Elsen, Rae,
  Vinyals, and Sifre}]{Hoffmann2022TrainingCL}
Jordan Hoffmann, Sebastian Borgeaud, Arthur Mensch, Elena Buchatskaya, Trevor
  Cai, Eliza Rutherford, Diego de~Las~Casas, Lisa~Anne Hendricks, Johannes
  Welbl, Aidan Clark, Tom Hennigan, Eric Noland, Katie Millican, George van~den
  Driessche, Bogdan Damoc, Aurelia Guy, Simon Osindero, Karen Simonyan, Erich
  Elsen, Jack~W. Rae, Oriol Vinyals, and L.~Sifre. 2022.
\newblock Training compute-optimal large language models.
\newblock \emph{ArXiv}, abs/2203.15556.

\bibitem[{Huebner et~al.(2021)Huebner, Sulem, Fisher, and
  Roth}]{Huebner2021BabyBERTaLM}
Philip~A. Huebner, Elior Sulem, Cynthia Fisher, and Dan Roth. 2021.
\newblock \href {https://api.semanticscholar.org/CorpusID:241583340}
  {Babyberta: Learning more grammar with small-scale child-directed language}.
\newblock In \emph{Conference on Computational Natural Language Learning}.

\bibitem[{Jiao et~al.(2019)Jiao, Yin, Shang, Jiang, Chen, Li, Wang, and
  Liu}]{Jiao2019TinyBERTDB}
Xiaoqi Jiao, Yichun Yin, Lifeng Shang, Xin Jiang, Xiao Chen, Linlin Li, Fang
  Wang, and Qun Liu. 2019.
\newblock Tinybert: Distilling bert for natural language understanding.
\newblock In \emph{Findings}.

\bibitem[{Lan et~al.(2019)Lan, Chen, Goodman, Gimpel, Sharma, and
  Soricut}]{Lan2019ALBERTAL}
Zhenzhong Lan, Mingda Chen, Sebastian Goodman, Kevin Gimpel, Piyush Sharma, and
  Radu Soricut. 2019.
\newblock Albert: A lite bert for self-supervised learning of language
  representations.
\newblock \emph{ArXiv}, abs/1909.11942.

\bibitem[{Liu et~al.(2019)Liu, Ott, Goyal, Du, Joshi, Chen, Levy, Lewis,
  Zettlemoyer, and Stoyanov}]{Liu2019RoBERTaAR}
Yinhan Liu, Myle Ott, Naman Goyal, Jingfei Du, Mandar Joshi, Danqi Chen, Omer
  Levy, Mike Lewis, Luke Zettlemoyer, and Veselin Stoyanov. 2019.
\newblock Roberta: A robustly optimized bert pretraining approach.
\newblock \emph{ArXiv}, abs/1907.11692.

\bibitem[{OpenAI(2023)}]{OpenAI2023GPT4TR}
OpenAI. 2023.
\newblock \href {https://api.semanticscholar.org/CorpusID:257532815} {Gpt-4
  technical report}.
\newblock \emph{ArXiv}, abs/2303.08774.

\bibitem[{Raffel et~al.(2019)Raffel, Shazeer, Roberts, Lee, Narang, Matena,
  Zhou, Li, and Liu}]{Raffel2019ExploringTL}
Colin Raffel, Noam~M. Shazeer, Adam Roberts, Katherine Lee, Sharan Narang,
  Michael Matena, Yanqi Zhou, Wei Li, and Peter~J. Liu. 2019.
\newblock Exploring the limits of transfer learning with a unified text-to-text
  transformer.
\newblock \emph{ArXiv}, abs/1910.10683.

\bibitem[{Sanh et~al.(2019)Sanh, Debut, Chaumond, and
  Wolf}]{Sanh2019DistilBERTAD}
Victor Sanh, Lysandre Debut, Julien Chaumond, and Thomas Wolf. 2019.
\newblock Distilbert, a distilled version of bert: smaller, faster, cheaper and
  lighter.
\newblock \emph{ArXiv}, abs/1910.01108.

\bibitem[{Vaswani et~al.(2017)Vaswani, Shazeer, Parmar, Uszkoreit, Jones,
  Gomez, Kaiser, and Polosukhin}]{Vaswani2017AttentionIA}
Ashish Vaswani, Noam~M. Shazeer, Niki Parmar, Jakob Uszkoreit, Llion Jones,
  Aidan~N. Gomez, Lukasz Kaiser, and Illia Polosukhin. 2017.
\newblock Attention is all you need.
\newblock In \emph{NIPS}.

\bibitem[{Wang et~al.(2019)Wang, Pruksachatkun, Nangia, Singh, Michael, Hill,
  Levy, and Bowman}]{Wang2019SuperGLUEAS}
Alex Wang, Yada Pruksachatkun, Nikita Nangia, Amanpreet Singh, Julian Michael,
  Felix Hill, Omer Levy, and Samuel~R. Bowman. 2019.
\newblock Superglue: A stickier benchmark for general-purpose language
  understanding systems.
\newblock \emph{ArXiv}, abs/1905.00537.

\bibitem[{Warstadt et~al.(2023)Warstadt, Mueller, Choshen, Wilcox, Zhuang,
  Ciro, Mosquera, Williams, Paranjabe, Linzen, and
  Cotterell}]{warstadt-et-al-2023-babylm}
Alex Warstadt, Aaron Mueller, Leshem Choshen, Ethan~Gotlieb Wilcox, Chengxu
  Zhuang, Juan Ciro, Rafael Mosquera, Adina Williams, Bhargavi Paranjabe, Tal
  Linzen, and Ryan Cotterell. 2023.
\newblock Findings of the 2023 {B}aby{LM} {C}hallenge: {S}ample-efficient
  pretraining on developmentally plausible corpora.
\newblock In \emph{Proceedings of the 2023 {B}aby{LM} {C}hallenge}. Association
  for Computational Linguistics (ACL).

\bibitem[{Warstadt et~al.(2019)Warstadt, Parrish, Liu, Mohananey, Peng, Wang,
  and Bowman}]{Warstadt2019BLiMPAB}
Alex Warstadt, Alicia Parrish, Haokun Liu, Anhad Mohananey, Wei Peng, Sheng-Fu
  Wang, and Samuel~R. Bowman. 2019.
\newblock Blimp: A benchmark of linguistic minimal pairs for english.
\newblock \emph{Transactions of the Association for Computational Linguistics},
  8:377--392.

\bibitem[{Warstadt et~al.(2020)Warstadt, Zhang, Li, Liu, and
  Bowman}]{Warstadt2020LearningWF}
Alex Warstadt, Yian Zhang, Haau-Sing Li, Haokun Liu, and Samuel~R. Bowman.
  2020.
\newblock \href {https://api.semanticscholar.org/CorpusID:222290865} {Learning
  which features matter: Roberta acquires a preference for linguistic
  generalizations (eventually)}.
\newblock In \emph{Conference on Empirical Methods in Natural Language
  Processing}.

\bibitem[{Wolf et~al.(2019)Wolf, Debut, Sanh, Chaumond, Delangue, Moi, Cistac,
  Rault, Louf, Funtowicz, and Brew}]{Wolf2019HuggingFacesTS}
Thomas Wolf, Lysandre Debut, Victor Sanh, Julien Chaumond, Clement Delangue,
  Anthony Moi, Pierric Cistac, Tim Rault, R{\'e}mi Louf, Morgan Funtowicz, and
  Jamie Brew. 2019.
\newblock Huggingface's transformers: State-of-the-art natural language
  processing.
\newblock \emph{ArXiv}, abs/1910.03771.

\bibitem[{Zhang et~al.(2022)Zhang, Roller, Goyal, Artetxe, Chen, Chen, Dewan,
  Diab, Li, Lin, Mihaylov, Ott, Shleifer, Shuster, Simig, Koura, Sridhar, Wang,
  and Zettlemoyer}]{Zhang2022OPTOP}
Susan Zhang, Stephen Roller, Naman Goyal, Mikel Artetxe, Moya Chen, Shuohui
  Chen, Christopher Dewan, Mona Diab, Xian Li, Xi~Victoria Lin, Todor Mihaylov,
  Myle Ott, Sam Shleifer, Kurt Shuster, Daniel Simig, Punit~Singh Koura, Anjali
  Sridhar, Tianlu Wang, and Luke Zettlemoyer. 2022.
\newblock Opt: Open pre-trained transformer language models.
\newblock \emph{ArXiv}, abs/2205.01068.

\end{thebibliography}
\bibliographystyle{acl_natbib}

\end{document}